\def\year{2020}\relax
\documentclass[letterpaper]{article} 
\usepackage{aaai19}  
\usepackage{times}  
\usepackage{helvet} 
\usepackage{courier}  
\usepackage[hyphens]{url}  
\usepackage{graphicx} 
\usepackage{color}
\usepackage{graphicx}  
\usepackage{amsmath}
\usepackage{multirow}
\usepackage{booktabs}
\usepackage{amssymb}
\newcommand{\entityHead}{\mathbf{h}}
\newcommand{\entityTail}{\mathbf{t}}
\newcommand{\relation}{\mathbf{r}}

\title{
Composing Knowledge Graph Embeddings via Word Embeddings\\ }
\author{Lianbo Ma \textsuperscript{\rm 1},Peng Sun\textsuperscript{\rm 1}\thanks{Corresponding authors: Peng Sun, 1871124@stu.neu.edu.cn, and Zhiwei Lin,  z.lin@ulster.ac.uk},Zhiwei Lin \textsuperscript{\rm 2} ,Hui Wang \textsuperscript{\rm 2}\\ 
\textsuperscript{\rm 1}College of Software, Northeastern University, Shenyang, China\\ 
\textsuperscript{\rm 2}School of Computing, Ulster University, Newtownabbey BT37 0QB, U.K\\
}
 
\begin{document}

\maketitle

\begin{abstract}

Learning knowledge graph embedding from an existing knowledge graph is very important to   knowledge graph completion. For a fact $(h,r,t)$ with the head entity $h$  having a relation $r$ with the tail entity $t$,   the current approaches aim to learn low dimensional representations $(\entityHead,\relation,\entityTail)$, 
each of which corresponds to  the elements in $(h, r, t)$, respectively.   As $(\entityHead,\relation,\entityTail)$ is learned from the existing facts within a knowledge graph, these representations can not be used to detect unknown facts (if the entities or relations never occur in the knowledge graph).

This paper proposes a new approach called TransW, aiming to go beyond the current work by composing knowledge graph embeddings using word embeddings.  Given the fact that an entity or a relation contains one or more words (quite often), it is sensible to learn a mapping function from word embedding spaces to knowledge embedding spaces, which shows  how entities are constructed using human words. More importantly, composing knowledge embeddings using word embeddings makes it possible to deal with the emerging new facts (either new entities or relations). Experimental results using three public datasets show the consistency and outperformance of the proposed TransW. 

\end{abstract}
\section{Introduction}
A {\em knowledge graph } $K=(E,R)$ where $E$ is a set of entities and $R$ is a set of relations,  contains  linked  information of  facts,  where each fact is a triple \footnote{This paper will   use fact and triple   interchangeably if without any confusion. } $(h,r,t)$  showing the relationship $r\in R$  between the head entity $h \in E$  and the tail entity $t \in E$. However, with the growing volume of data on the Internet, the new facts emerge constantly and they need to be added into the existing knowledge graphs in order to complete the graphs. As such, knowledge graphs are far from complete. Adding the new facts manually is labor intensive and also makes it difficult to validate if the new fact should belong to the knowledge graph. One way around this is to  learn {\em knowledge graph embeddings} for the entities and their relations (i.e., encoding both entities and their relations within facts into a continuous low-dimensional vector space) \cite{DBLP:conf/nips/BordesUGWY13,tkde:survey,Trouillon:2017:JMLR,DBLP:conf/aaai/WangLP18}. For a triple $(h,r,t)$, let $(\entityHead,\relation,\entityTail)$ be its representation in knowledge graph embedding, where $\entityHead, \entityTail, \relation \in \mathbb{R}^n$. The existing approaches aim to train a model to  `translate' the head $\entityHead$ to the tail entity $\entityTail$ using the relation $\relation$ with minimum loss, such as  $\mathbf{h + r \approx t}$ in the   TransE \cite{DBLP:conf/nips/BordesUGWY13} model.

\begin{figure*}[!h]
	\begin{center}
		\centering
		\includegraphics[width=140mm]{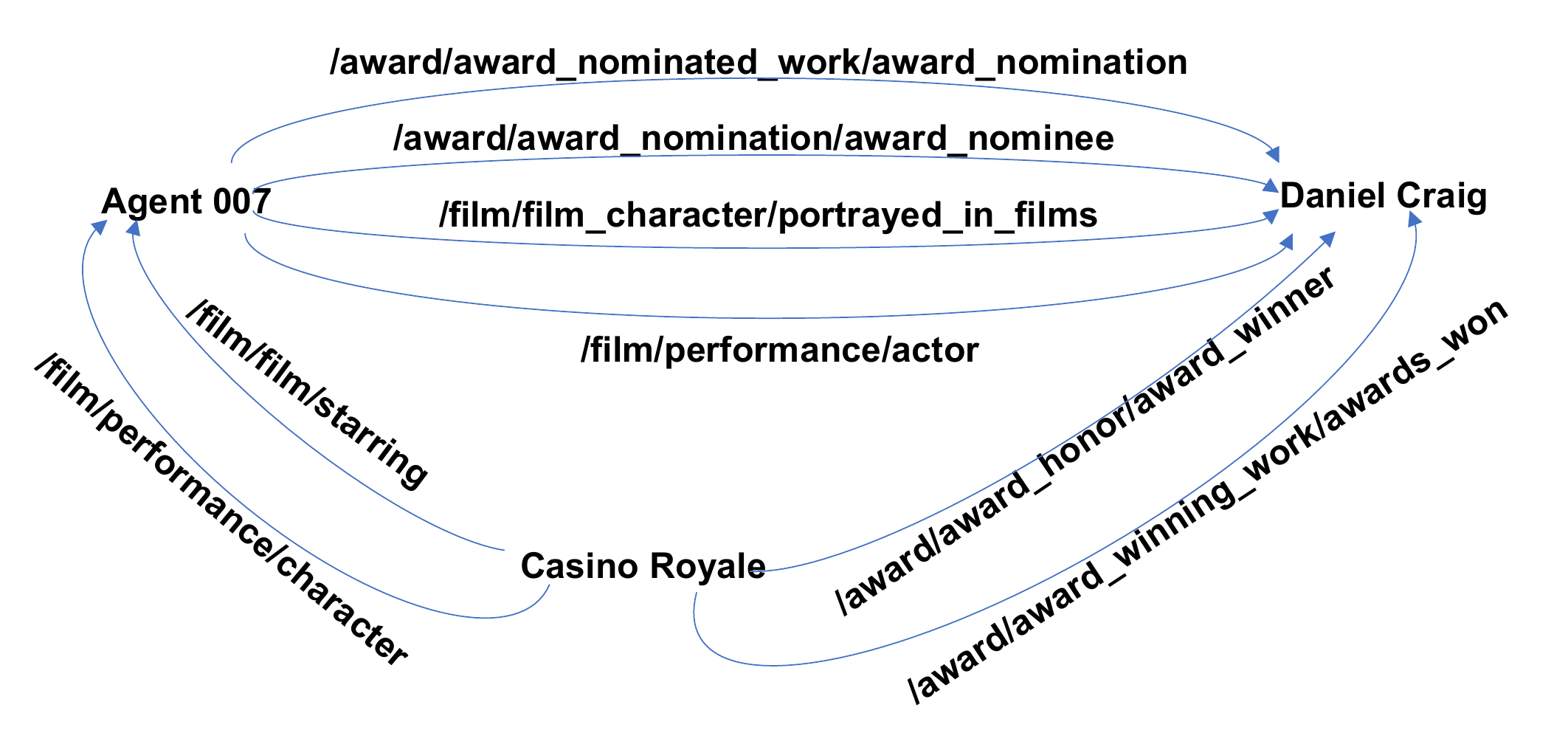}\\
	\end{center}
	\caption{An example of a sub-graph of the knowledge graph from FB15K dataset, where multiple relations exist between two entities}
	\label{figure:daniel:craig:example}
\end{figure*}


However, as the   representations of $(\entityHead,\relation,\entityTail)$  in knowledge graph embeddings are trained with the known entities from $E$ and known relations from $R$,  the learned representations  have critical issues for dealing with `unknown' new entities and new relations (i.e,   a new fact $(h,r,t)$ where $h\notin E$ or $t\notin E$ or $r\notin R$). For example,  Fig. \ref{figure:daniel:craig:example} shows a sub-graph of the knowledge graph extracted from FB15K dataset~\cite{DBLP:conf/nips/BordesUGWY13} , in which there are 4 relations between ``Agent 007'' and ``Daniel Craig''.  There may be more than 4 relations between the two entities and the new relations may not exist in $R$. Also, the title of the new ``007'' film (beyond  2020) is still  new, and it would not exist in $E$. Therefore,  the current approaches can not deal with the unseen entities and relations.

With the fast growing volume of data on the Internet,  adding new entities and relations is very important in  scaling out   knowledge graphs for enrichment. The current knowledge graph embedding approaches learn $(\entityHead,\relation,\entityTail)$ for a specific fact $(h,r,t)$ by ignoring the detail of the words within the facts. Such limitation makes them very difficult to generalize to unknown facts.
This paper presents a new approach called TransW, which aims to address these issues by learning knowledge graph embeddings via the composition of word embeddings due to the fact that each entity or relation may contain multiple words. To the best of our knowledge, this is the first work aiming to enrich a knowledge graph by detecting unknown entities and unknown relations, using word embeddings.

From Fig. \ref{figure:daniel:craig:example}, it is very obvious that the multiple relations between two entities are very similar. For example, ``/film/film/staring'' and ``/film/performance/character'' are very similar as (1) on one hand, they all share the keyword ``film'', and (2) on the other hand, ``staring '' is very close to ``performance/character''. Each word can be represented in word embeddings and the words with semantic similarity tend to have very high similarity between their word embeddings.  The similarity between two entities (or two relations) is measurable if  a mapping function from word embedding to knowledge graph embedding can be defined. Let $\mathbb{W}$ be the word embedding space and $\mathbb{K}$ be the knowledge embedding space, and $g: \mathbb{W} \rightarrow \mathbb{K}$ be the mapping function from word embedding space to knowledge embedding space,  the aim of this work is to learn such mapping function, so that for any new entities or new relations, their representations in knowledge graph embeddings can be constructed via this mapping function. This will be much flexible in dealing with the unknown entities and relations. 

The contributions of this paper include (1) proposing to use word embeddings as the ingredients for learning knowledge graph embedding; (2)  introducing a new approach called TransW in order to create mappings from word embeddings to knowledge graph embeddings; (3) conducting experiments to show the proposed approach can deal with unknown facts.

\section{Related Work}
\newcommand{\tabincell}[2]{\begin{tabular}{@{}#1@{}}#2\end{tabular}}
\begin{table*}[!h]
	\centering
	\caption{Different translation-based models: the scoring functions $f_{r}(\mathbf{h}, \mathbf{t})$ and the number of parameters. ${n_e}$ and ${n_r}$ are the number of unique entities and relations, respectively. ${k}$  is the dimension of embedding space. $n$ and ${m}$ are the number of words of each entitiy and relation}
	\begin{tabular}{ |c | c | c| }   \hline
		Model & Score function $f_{r}(\mathbf{h}, \mathbf{t})$ & Parameters\\ \hline
		
		TransE  & $\|\mathbf{h}+\mathbf{r}-\mathbf{t}\|_{\ell_{1 / 2}}^{2}$ \quad $\mathbf{r} \in \mathbb{R}^{k}$ & ${O(n_e k + n_r k)}$
		\\
		TransH & $\left\|\left(\mathbf{h}-\mathbf{w}_{r}^{\top} \mathbf{h} \mathbf{w}_{r}\right)+\mathbf{d}_{r}-\left(\mathbf{t}-\mathbf{w}_{r}^{\top} \mathbf{t} \mathbf{w}_{r}\right)\right\|_{\ell_{1 / 2}}^{2}$
		\quad$\mathbf{r}, \mathbf{w}_{r} \in \mathbb{R}^{k}$
		& $O\left(n_{e} k+2 n_{r} k\right)$
		\\
		TransR  & $\left\|\mathbf{M}_{r} \mathbf{h}+\mathbf{r}-\mathbf{M}_{r} \mathbf{t}\right\|_{\ell_{1 / 2}}^{2}$ \quad $\mathbf{r} \in \mathbb{R}^{k}, \mathbf{M}_{r} \in \mathbb{R}^{k \times k}$
		& $O\left(n_{e} k+n_{r}(k+1) k\right)$
		\\
		TransD  &$\left\|\left(\mathbf{w}_{r} \mathbf{w}_{h}^{\top}+\mathbf{I}\right) \mathbf{h}+\mathbf{r}-\left(\mathbf{w}_{r} \mathbf{w}_{t}^{\top}+\mathbf{I}\right) \mathbf{t}\right\|_{\ell_{1 / 2}}^{2}$ \quad $\mathbf{r}, \mathbf{w}_{r} \in \mathbb{R}^{k}$
		&$O\left(2 n_{e} k+2 n_{r} k\right)$
		\\
		TranSparse &
		$\left\|\mathbf{M}_{r}\left(\theta_{r}\right) \mathbf{h}+\mathbf{r}-\mathbf{M}_{r}\left(\theta_{r}\right) \mathbf{t}\right\|_{\ell_{1 / 2}}^{2}$\quad
			$\mathbf{r} \in \mathbb{R}^{k}, \mathbf{M}_{r}\left(\theta_{r}\right) \in \mathbb{R}^{k \times k}$
		& $O\left(n_{e} k+n_{r}(k+1) k\right)$
		\\
		\hline
		TransW  &
		
			$\|(\sum \mathbf{h}_{{i}} \otimes \mathbf{w}_{{h i}}+\mathbf{b}_{{h}})+\sum\mathbf{r}_{i} \otimes \mathbf{w}_{{r i}}$ $-(\sum \mathbf{t}_{{i}} \otimes \mathbf{w}_{{t i}} +\mathbf{b}_{{t}}) \|_{\ell_{1 / 2}}^{2}$\quad$ \mathbf{w}_{{r i}} \in \mathbb{R}^{k}$
		& ${O((n+1) n_e k + (m+1) n_r k)}$
		\\
		\hline
	\end{tabular}
	\label {Parameters}
\end{table*}
\subsection{Translation-based Models}
For   a   triplet ($h$, $r$, $t$),  the \textbf{TransE} \cite{DBLP:conf/nips/BordesUGWY13} treats the relation $r$ as a translation from a head entity $h$ to a tail entity $t$ . Here, ($\mathbf{h} + \mathbf{r}$) is close to ($\mathbf{t}$) and the score function is 
\begin{equation}\label{eq:transe}
f_{r}(\mathbf{h},\mathbf{t})=-\|\mathbf{h}+\mathbf{r}-\mathbf{t}\|_{\ell_{1 / 2}}^{2}.
\end{equation}
Despite its success in knowledge graphs with many  1-1 relations, it is not suitable for $1-N$, $N-1$ and $N-M$ relations. To address these issues, the  \textbf{TransH} \cite{DBLP:conf/aaai/WangZFC14} transforms the entities using a norm vector $\mathbf{w_r}$:
\begin{eqnarray}
 \mathbf{h_{\perp}}=\mathbf{h}-\mathbf{w_{r}^{\top}} \mathbf{h w_{r}}\\\mathbf{t_{\perp}}=\mathbf{t}-\mathbf{w_{r}^{\top} t w_{r}}. 
\end{eqnarray}
 before using Eq. \eqref{eq:transe}.
 
The  \textbf{TransR} \cite{DBLP:conf/aaai/LinLSLZ15}  defines a transformation matrix $\mathbf{M_{r}}$  and its score function as 
\begin{equation}
\left\|\mathbf{M}_{r} \mathbf{h}+\mathbf{r}-\mathbf{M}_{r} \mathbf{t}\right\|_{\ell_{1 / 2}}^{2}   
\end{equation}
The  \textbf{TransD} \cite{DBLP:conf/acl/JiHXL015} constructs a dynamic mapping matrix for each entity-relation pair. \textbf{TranSparse} \cite{DBLP:conf/aaai/JiLH016} uses sparse matrices to model the relations. Similar approaches also involve \textbf{TransM} \cite{DBLP:conf/paclic/FanZCZ14}, \textbf{TransG} \cite{DBLP:journals/corr/0005HHZ15} and \textbf{ManifoldE} \cite{DBLP:conf/ijcai/0005HZ16}. Recently, some new methods aim to deal with extra information, such as relation-type \cite{DBLP:conf/ijcai/WangWG15}, paths with different confidence levels (\textbf{PTransE}) \cite{PTransE}, data uncertainty (\textbf{KG2E}) \cite{DBLP:conf/cikm/HeLJ015}, and semantic smoothness of embedding space \cite{DBLP:journals/tkde/GuoWWWG17}. Besides, several recent TransE variants such as \textbf{TorusE} \cite{DBLP:conf/aaai/EbisuI18} and \textbf{ConvKB} \cite{DBLP:conf/naacl/NguyenNNP18} also achieve the state-of-the-art performance.
\subsection{Other Models}
The structured embedding    \cite{DBLP:conf/aaai/BordesWCB11} uses two relation-specific matrices ($\mathbf{M_{h,r}}$, and $\mathbf{M_{t,r}}$) for projecting head and tail entities, respectively, and defines a new score function as  $f_{r}(\mathbf{h, t})=\|\mathbf{M_{h,r} h-M_{t,r} t}\|$. Unstructured Model (\textbf{UM}) model \cite{DBLP:journals/jmlr/BordesGWB12} is a naive version of TransE, where relation information is ignored and the score function is reduced to $f_{r}(\mathbf{h, t})=\|\mathbf{h-t}\|_2^{2}$. Single Layer Model (\textbf{SLM}) \cite{DBLP:conf/nips/SocherCMN13} constructs a nonlinear neural network to represent the score function, which is defined as $f_{r}(\mathbf{h, t})=\mathbf{u_r^{\top}}g(\mathbf{M_{r1} h}+\mathbf{M_{r2} t})$ where $\mathbf{M_{r1}}$ and $\mathbf{M_{r2}}$ are relation-specific weight matrices.  Semantic Matching Energy (\textbf{SME})  \cite{DBLP:journals/corr/abs-1301-3485} aims to capture correlations between entities and relations via multiple matrix products and Hadamard product, defined as $f_{r}(\mathbf{h, t})=(\mathbf{M_1 h}+\mathbf{M_2 r}+ \mathbf{b_1})^{\top}(\mathbf{M_3 t}+\mathbf{M_4 r}+ \mathbf{b_2})$ and $f_{r}(\mathbf{h, t})=(\mathbf{M_1 h}\otimes \mathbf{M_2 r}+ \mathbf{b_1})^{\top}(\mathbf{M_3 t}\otimes \mathbf{M_4 r}+ \mathbf{b_2})$, where $\mathbf{M_1}$, $\mathbf{M_2}$, $\mathbf{M_3}$ and $\mathbf{M_4}$ are weight matrices, $\otimes$ is the Hadamard product, $\mathbf{b_1}$ and $\mathbf{b_2}$ are bias vectors. Latent Factor Model (\textbf{LFM}) \cite{DBLP:conf/nips/JenattonRBO12}  incorporates second-order correlations between entities using a quadratic form, and defines a bilinear score function as $f_{r}(\mathbf{h, t})=\mathbf{h^{\top}W_r t}$. Neural Tensor Network (\textbf{NTN}) \cite{DBLP:conf/nips/SocherCMN13} defines an expressive score function to extend the SLM model, i.e, $f_{r}(\mathbf{h, t})=\mathbf{u_r^{\top}} \tanh(\mathbf{h^{\top} W_r t}+\mathbf{W_{r,1} h}+\mathbf{W_{r,2} t}+\mathbf{b_r})$, where  $\mathbf{u_r}$ is a relation-specific linear layer,  
$\mathbf{W}\in {\mathbb{R}}^{d \times d \times d \times k}$ is a 3-way tensor,
In addition of the above approaches, we will also compare with another common model \textbf{RESCAL} in the experiments, which is a collective matrix factorization model \cite{DBLP:conf/icml/NickelTK11}. More recently,  adversarial learning has also been investigated for learning knowledge graph embeddings \cite{KBGAN:2018,DBLP:conf/aaai/WangLP18}.  

\section{The TransW Approach}
As  the current knowledge embedding approaches ignore the fine-grained semantic information in the word space, and may be insufficient to deal with new facts with  unknown entities or relations, this section presents the TransW approach     which models entities and relations via word embeddings. 

\subsection{A motivating example}
{

To begin with, using Fig. \ref{figure:daniel:craig:example} as an example graph, the curret approach to learn $(\entityHead,\relation,\entityTail)$ for the triple (``Agent 007'', ``/film/film/starring'', ``Casino Royale''), does not have semantic knowledge of how entity ``Agent 007'' is linked with ``Casino Royale''. The vector  $\entityHead$, $\relation$, and  $\entityTail$ are only the vector representations without any meaning. This makes it unhelpful in predicting ``/film/performance/character'' as a relation between ``Agent 007''  and  ``Casino Royale'', even if ``/film/performance/character'' is semantically related to  ``/film/film/starring''.

Suppose in the word embedding space, if  the embeddings for  ``film'', ``performance'', ``character'' and ``starring'' are $\entityHead_f$, $\entityHead_p$, $\entityHead_c$ and $\entityHead_s$, by using a linear mapping function $g$ to combine the word embeddings for the relations:
$$g(\entityHead_f,\entityHead_s) \text{\hspace{0.2cm}    for ``/film/film/starring''}$$   
$$g(\entityHead_f,\entityHead_p,\entityHead_c) \text{\hspace{0.2cm}    for ``/film/film/starring''}$$   
it would be much easier to show the similarity between two relations in knowledge graph spaces. This will also make it possible to detect unknown facts in order to scale out a knowledge graph for enrichment.

\subsection{TransW}
 
In TransW, each entity or relation is represented in the form of linear combination of word embeddings.  For a triple $(h,r,t)$ and its embedding ($\mathbf h$, $\mathbf r$, $\mathbf t$), suppose the numbers of words in $h$, $r$ and $t$ are  $n$, $p$ and $m$    respectively, then ($\mathbf h$, $\mathbf r$, $\mathbf t$) can be represented with their words embeddings:
%
\begin{equation}
\begin{split}
\mathbf{h}&=\sum_{i=0}^{\mathrm{n}} \mathbf{h}_{i}\otimes\mathbf{w}_{{hi}}+\mathbf{b}_{ {h}} \\
\mathbf{t}&=\sum_{i=0}^{\mathrm{m}} \mathbf{t}_{i}\otimes \mathbf w_{ {ti}}+\mathbf{b}_{ {t}} \\
\mathbf{r}&=\sum_{i=0}^{\mathrm{p}} \mathbf{r}_{i}\otimes \mathbf w_{ {ri}}+\mathbf{b}_{ {r}}
\label{TransW-2}
\end{split}
\end{equation}
\noindent where $\mathbf h_i$, $\mathbf r_i$   $\mathbf t_i \in \mathbb{W}$ are the word embedding for the $i$-th  word  in the corresponding $ h$, $  r$ and $  t$, respectively, ${\otimes}$ denotes Hadamard product, $\mathbf w_{  hi}$, $\mathbf w_{  ri}$, and $\mathbf w_{  ti}$ are the $i$-th connection vector for $\mathbf h$, $\mathbf r$ and $\mathbf t$, respectively, $\mathbf{b}_{ {h}}$, $\mathbf{b}_{ {t}}, \mathbf{b}_{ {r}}  \in \mathbb{R}^{k}$ are the bias parameters for entities $\mathbf{h}$ and $\mathbf{t}$,  and relation $\relation$ respectively.


Similar to that of TransE, Eq. \eqref{eq:transe} is used as  the score function  for TransE. 
The score is expected to be lower for a golden triplet and higher for an incorrect triplet. Let $\Delta$ be the set of all triplets, which are valid and $\Delta^\prime$ be the set of all triplets which are not valid, the loss function is defined according to the translation approach:

\begin{equation}
L=\sum_{\xi \in \Delta} \sum_{\xi^{\prime} \in \Delta^{\prime}}\left[\gamma+f_{r}\left(\xi^{\prime}\right)-f_{r}(\xi)\right]_{+}
\label{loss}
\end{equation}
where $\gamma$ is margin.

\textbf {Remark}: From Eq. (\ref{TransW-2}), it is obvious that each word embedding is assigned with a unique connection vector and an entity or relation is the sum of all transformed word embeddings. In this way, TransW represents each entity or relation by a unique combination of word embeddings. As illustrated in Fig.\ref{Illustration of TransW-2}, given the fact about ``\emph{/film/film/staring}'', TransW is able to predict a new relation ``\emph{/film/performance/character}'' according to the word-level semantic similarity between the two relations.

\begin{figure}

	\centering
	\includegraphics[width=90mm]{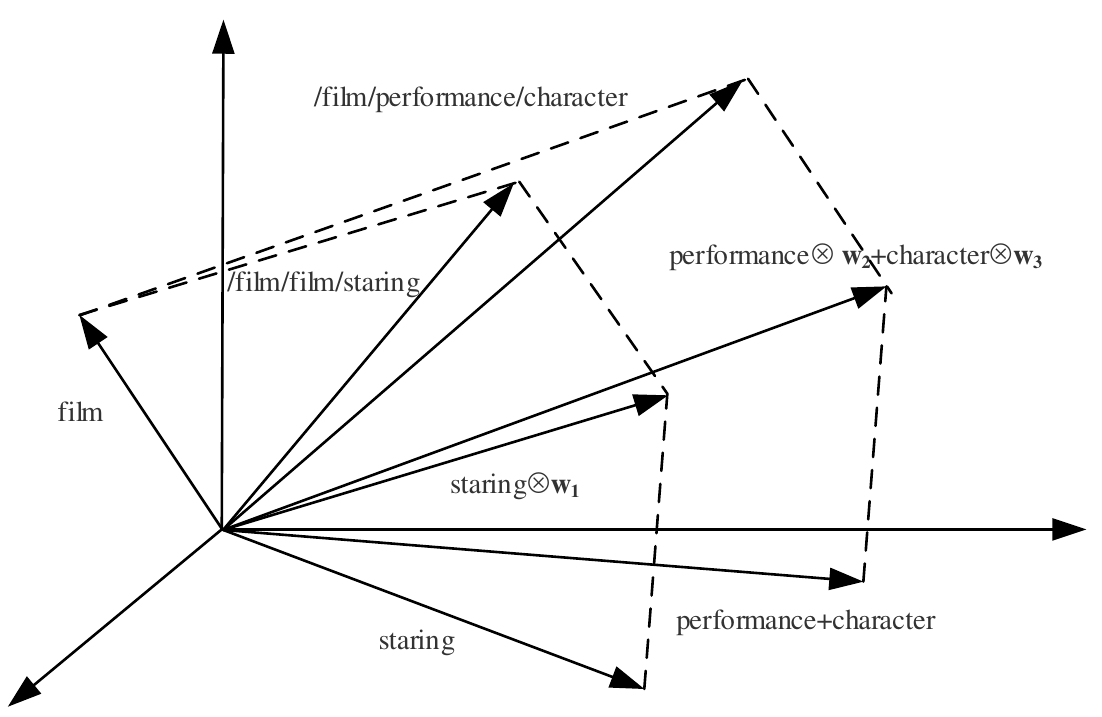}\\
	\caption{Illustration of composing relations using word embeddings by TransW}
	\label{Illustration of TransW-2}
\end{figure}
\section{ Experiments}
{ 

This paper uses  word embeddings in GloVe word embedding glove.6B.100d
\cite{pennington2014glove}. As this paper assumes the dimension of knowledge graph embeddings is 100, the word embeddings with dimension of 100 are used.   The learning rate is 0.01.

Three public benchmark datasets, as summarized in Table \ref{table:dataset}, are used to evaluate both TransW. Usually, the knowledge graph embeddings are evaluated using link prediction and triple classification. In link predication and triple classification, they assume the predication is based on the pre-defined set of entities $E$ and the pre-defined set of  relations  $R$. The existing work, such as TransE, etc, can not be used to decide if a new fact $(h,r,t)$ belongs to a knowledge graph when the new fact contains either an unseen entity $h\notin E$ or $t\notin E$ or an unseen relation $r\notin R$. Therefore, in addition to link prediction and triple classification, this paper sets out a new task for detecting unknown facts  with ``new relations''.

As such, this section contains three tasks for evaluating the proposed TransW:
\begin{enumerate}
	\item Detecting unknown facts: the relations in the test dataset  are unknown (do not occur in the training part);
	\item Link prediction: the relations and entities in the test dataset already exist in the training set. 
	\item Triple classification: the relations and entities in the test dataset already exist in the training set. 
\end{enumerate}

\begin{table}[t]
	\centering
	\caption{Summary of the public datesets used in the experiments.  Rel/Ent/Train/Valid/Test are the number of relations/entities/traing triples/validation triples/testing triples. }
	\begin{tabular}{|c|c c c c c|}
		\hline
		Dataset & Rel & Ent & Train & Valid & Test \\  \hline
		WN11 & 11 & 38,696 & 112,581 & 2,609 & 10,544 \\
		FB13 & 13 & 75,043 & 316,232 & 5,908 & 23,733 \\
		FB15K & 1345 & 14951 & 483142 & 50000 & 59071 \\  \hline
	\end{tabular}
	\label{table:dataset}
\end{table}

\subsection{Detecting Unknown Facts}

\begin{table*}[!t]
	\centering 
	\caption{Prediction accuracy for  unseen  facts of FB15K using 10-fold cross validation.}
	\begin{tabular}{|c|c c c c c |}   \hline 
		Fold  & Train(facts/relations) &Test(facts/relations) &Average (\%) &Bias(\%)& $\sigma$\\ \hline
		1 &444422/1210&77436/135 &61.18 &(-1.9, +1.14) &0.0791
		\\ 
		2 &437977/1211&10942/134 &54.62  &(-1.22,+0.78 ) &0.0985
		\\ 
		3 &455313/1210&50144/135 &55.07 &(-0.53 ,+0.53)  &0.0705
		\\ 
		4 &424743/1210&14430/135 &54.54 & (-1.42,+0)  &0.0768
		\\ 
		5 &416196/1210&16322/135 &52.4 &(-1.5,+1.0)  &0.0961
		\\ 
		6 &429286/1211&12952/134&55.76 &(-0.14,+0.29) &0.0993
		\\ 
		7 &404590/1210&156120/135 &53.82 &(-1.12,+0.84)  &0.0806
		\\ 
		8 &447139/1211& 8652/134&55.63 &(-0.23,+0.27 )  &0.0723
		\\ 
		9 &442557/1211&9955/134 &55.27 &(-0.19 ,+0.45) &0.1057
		\\ 
		10&446054/1211&9168/134 &56.69 &(-0.19 ,+0.41) &0.0845
		\\ \hline
		Average&&  &55.5 &(-0.84,+0.48) &  
		\\ 
		
		\hline 
	\end{tabular} 
	\label{table:10-fold}
\end{table*}  

The first part of the experiments carried out is to detect unknown facts.  Different from the existing approaches without using word embeddings, the TransW approach composes knowledge graph embeddings $(\entityHead,\relation, \entityTail)$ using word embeddings, which makes it possible to detect new facts $(h,r,t)$, where ``unseen'' entities ($h\notin E$ or $t\notin E$) or an ``unseen'' relation ($r\notin R$) exists. In this experiment, the primary focus will be evaluating the detection of  new relations using FB15K dataset. The reason of not using WN11 and FB13 is that they have very limited number of relations (as shown in Table \ref{table:dataset}, there are only 11 relations in WN11 and 13 relations in FB13), which makes it hard to train a model using just 10 relations in order to make accurate predication for the remaining relations (if 10-fold cross validation is used).

This evaluation is conducted using 10-fold cross validation and the 1345 relations in FB15K  are split  into 10 folds, from which 9 folds are used to train a model  and leave the  relations in the remaining  fold as ``unseen'' relations. This setting will make sure that the unseen facts will not only contain ``unseen'' relations ($r\notin R$) but  may also contain ``unseen'' entities. The hypothesis is that the TransW models are able to tell if the ``unseen'' facts are part of the existing knowledge graph or not.

Table \ref{table:10-fold} shows the statistical information of each fold. For example, in the 1st fold, there are 1210 relations and 444,422 facts in total for training . There are also 77436 facts and 135 relations in the test set. However, as the numbers of facts for the relations in the test set  are imbalanced (e.g., some relations in the test set only contains very few facts),  for each relation, only 5000 facts are randomly selected for testing. This is   repeated for 10 times in order to make sure if TransW can provide consistent results. The accuracy in the table is based on the average of the accuracies of these 10 times of testing subsets.

After training, a threshold $\sigma$ for determining if a relation is valid or not is set according to the score function of Eq. \eqref{eq:transe} using the training set in each fold. If the score (using Eq. \eqref{eq:transe}) for a fact in the testing set of each fold is lower than $\sigma$, this fact is regarded as a true fact and otherwise, the fact is not a valid fact to the knowledge graph.  The threshold $\sigma$ is shown in Table \ref{table:10-fold} for each fold, which is around 0.0863. 

From Table \ref{table:10-fold}, it is found that Trans-W performs very consistently in each fold, and the average accuracy for detecting unknown facts is very satisfying.

\begin{table*}[!h]
	\centering
	\caption{ Results on FB13\&WN11 for link prediction (\%)}
	\begin{tabular}{|c | cp{0.66cm} | cp{0.68cm} |cp{0.66cm} | cp{0.68cm} |cp{0.65cm}| cp{0.66cm} | }   \hline
		& \multicolumn{6}{c|}{FB13}  &\multicolumn{6}{c|}{WN11} \\ \hline
		&\multicolumn{2}{c|}{HIT@10}  &\multicolumn{2}{c|}{HIT@3} &\multicolumn{2}{c|}{HIT@1} & \multicolumn{2}{c|}{HIT@10}   &\multicolumn{2}{c|}{HIT@3} &\multicolumn{2}{c|}{HIT@1}\\
		Metric & Raw & Filter & Raw & Filter & Raw & Filter & Raw & Filter & Raw & Filter & Raw & Filter\\
		\hline
		Rescal &23.86 &24.59 &18.86 &19.80&5.70 &6.92 &2.64 &2.84 &1.52  &1.65  &0.81  &0.97\\
		TransE &32.65 &32.75 &24.86 &25.16 &17.28 &17.57  &25.84 &30.75	&15.37	&19.47	&7.2 &10.23\\
		TransR &26.83 &27.53 &20.34 &20.94 &14.04 &14.97 &8.38 &8.66 &4.72 &4.84 &2.02 &2.24\\
		TransH &25.02 &25.80&18.03 &18.78 &6.62 &7.52 &23.87 &29.62	&13.17	&19.13	&4.97	&9.29  \\
		TransD &34.77&35.22 &26.75 &27.77 &18.47 &19.85 &25.99 &26.73 &19.14 &19.84 &7.99	& 8.84\\
		DistMult & 16.72 & 17.82& 14.12 & 14.80 & 6.27 & 7.03  &-	&-	&-	&-	&-&	-
		\\
		Complex & 16.51 & 17.60 & 14.21 & 14.80& 6.80 & 7.53 &14.68&15.94 &10.06 &10.99 &6.3 &7.4\\
		\hline
		TransW &\textbf{45.12} &\textbf{47.93}&\textbf{35.71}&\textbf{35.94}&\textbf{29.77} &\textbf{30.62} &21.15 &23.2 &13.35 &14.26 &4.27 &5.23\\
		\hline
	\end{tabular}
	\label{table:result:link:prediction}
\end{table*}

\begin{table*}[!h]
	\centering
	\caption{Accuracy for triple classification on FB13 and WN11}
	\begin{tabular}{|c|c|c|}
		\hline
		Model & FB13 (\%)  & WN11 (\%)   \\ \hline
		SE ~\cite{DBLP:conf/aaai/BordesWCB11}& 75.2 & 53.0 \\
		SME~\cite{DBLP:journals/corr/abs-1301-3485} &63.7 & 70.0 \\
		SLM~\cite{DBLP:conf/nips/SocherCMN13}&85.3  & 69.9 \\
		LFM ~\cite{DBLP:conf/nips/JenattonRBO12}&84.3 & 73.8\\
		NTN~\cite{DBLP:conf/nips/SocherCMN13}&87.1 & 70.4 \\
		Rescal ~\cite{DBLP:conf/icml/NickelTK11}& 70.7 &61.0\\
		TransE ~\cite{DBLP:conf/nips/BordesUGWY13} & 78.2 & 75.7  \\
		TransH ~\cite{DBLP:conf/aaai/WangZFC14}& 77.4  & 77.7 \\
		TransR ~\cite{DBLP:conf/aaai/LinLSLZ15}& 82.5  & 85.5\\
		TransD ~\cite{DBLP:conf/acl/JiHXL015}&\textbf{87.7} & 86.4\\
		DistMult ~\cite{DBLP:journals/corr/YangYHGD14a}&55.3 & 50.0 \\
		Complex ~\cite{DBLP:conf/icml/TrouillonWRGB16}&56.2 & 60.0  \\
		TranSparse ~\cite{DBLP:conf/aaai/JiLH016}& 86.7 & 86.3 \\
		ManifoldE ~\cite{DBLP:conf/ijcai/0005HZ16}& 87.2 &\textbf{87.5} \\
		\hline
		TransW (this paper) & \textbf{87.5} &81.1\\
		\hline
	\end{tabular}
	\label{table:triple:classification}
\end{table*}
}
\subsection  {Link Prediction}
Link prediction is to  locate  entities from $E$ for the following two cases:
\begin{enumerate}
	\item $(?, r,t)$ when the   head entity is missing;
	\item $(h,r,?)$ when  the   tail entity is missing;
\end{enumerate}
where $h,t\in E$ and $r\in R$, in order to complete a triplet.

This experiment follows the settings in {  \cite{DBLP:conf/aaai/BordesWCB11,DBLP:conf/nips/BordesUGWY13}}, using  WN11 and FB13 datasets.

In order to carry out the experiments, for each triple $(h,r,t)$ in the test set, a new set $T$ of data  is created:
$$T=\{ (e,r,t) | \forall e\in E\}\cup \{ (h,r,e) | \forall e\in E \}. $$
which contains ``true'' triples and ``false'' triples.

Scores are calculated according to the score function in Eq. \eqref{eq:transe} for the ``true'' and ``false'' triples in $T$. The triples related to $(?, r,t)$ will be ranked in descending order based on their scores. This is repeated for  $(h,r,?)$.

To compare with the existing approaches, this paper uses the \emph {HITS@N} approach, which calculates the proportion of correct predictions in the top $N$ facts from the ranking lists. This means, the higher \emph {HITS@N}, the better performance it has.
In Table \ref{table:result:link:prediction}, this paper reports the results for $HITS@10$, $HITS@3$ and $HITS@1$ using FB13 and WN11 datasets.

From Table \ref{table:result:link:prediction},  TransW  is significantly better than  the state of the art approaches   for FB13 dataset. Similar to FB15K, the FB13 comes from Freebase. As shown in Fig. \ref{figure:daniel:craig:example}, the relations between entities has very rich context with multiple words. The composition of knowledge graph embeddings using word embeddings makes use of the semantic information in word embeddings and as such the performance is significantly improved.    

Whilst in WN11 dataset, which is from the WordNet, each entity in WN11 is a word and the relations between entities do not have rich semantic information. Therefore, the results are not as good as what TransW achieves in FB13.

\subsection {Triple Classification}

  Triple classification is to evaluate if a triple $(h,r,t)$ is a correct fact in a knowledge graph. False facts are also added to the 3 datasets, according to the settings in \cite{DBLP:conf/nips/SocherCMN13} and \cite{DBLP:conf/aaai/WangZFC14}. Therefore, this triple classification is a binary classification with both ``true'' and ``false'' facts. 
 
A relation-specific threshold $\sigma_r$, is chosen based on the best classification accuracies on the validation dataset.
For a triple ($ h$, $ r$, $ t$), if its score of $fr(\mathbf h, \mathbf t)\leq \sigma_r$, then ($ h$, $ r$, $ t$) is  classified as positive, otherwise negative.

The result for TransW in FB13 is better than most of the existing approaches though it is 0.2\% lower than TransD. TransW has a satisfying accuracy compared to the original TransE in WN11. The results for triple classification confirms the finding in the experiment for  link predication that utilizing the semantic information would help to enrich knowledge graph embedding.

%

\section {Conclusions and Future Work}
This paper introduces a new approach, called TransW, which encodes  entities and relations into a low dimentional space via word embeddings.  Using such composition with word embeddings, the word-level semantic information hidden in the word embedding spaces can be utilized for detecting  unknown facts, where entities or relations never occur in the existing knowledge graphs. Experimental results using public datasets validate the hypothesis. Future work includes  1) defining a better representation for  entities and relations via word embeddings; 2) testing the approach using various score functions in order to improve the accuracy for detecting unknown facts. 

\bibliographystyle{apalike}  
\bibliography{aaai.bib}

\begin{thebibliography}{}

\bibitem[\protect\citeauthoryear{Bordes \bgroup et al\mbox.\egroup
  }{2011}]{DBLP:conf/aaai/BordesWCB11}
Bordes, A.; Weston, J.; Collobert, R.; and Bengio, Y.
\newblock 2011.
\newblock Learning structured embeddings of knowledge bases.
\newblock In Burgard, W., and Roth, D., eds., {\em Proceedings of the
  Twenty-Fifth {AAAI} Conference on Artificial Intelligence, {AAAI} 2011, San
  Francisco, California, USA, August 7-11, 2011}.
\newblock {AAAI} Press.

\bibitem[\protect\citeauthoryear{Bordes \bgroup et al\mbox.\egroup
  }{2012}]{DBLP:journals/jmlr/BordesGWB12}
Bordes, A.; Glorot, X.; Weston, J.; and Bengio, Y.
\newblock 2012.
\newblock Joint learning of words and meaning representations for open-text
  semantic parsing.
\newblock In Lawrence, N.~D., and Girolami, M.~A., eds., {\em Proceedings of
  the Fifteenth International Conference on Artificial Intelligence and
  Statistics, {AISTATS} 2012, La Palma, Canary Islands, Spain, April 21-23,
  2012}, volume~22 of {\em {JMLR} Proceedings},  127--135.
\newblock JMLR.org.

\bibitem[\protect\citeauthoryear{Bordes \bgroup et al\mbox.\egroup
  }{2013}]{DBLP:conf/nips/BordesUGWY13}
Bordes, A.; Usunier, N.; Garc{\'{\i}}a{-}Dur{\'{a}}n, A.; Weston, J.; and
  Yakhnenko, O.
\newblock 2013.
\newblock Translating embeddings for modeling multi-relational data.
\newblock In Burges et~al. \shortcite{DBLP:conf/nips/2013},  2787--2795.

\bibitem[\protect\citeauthoryear{Burges \bgroup et al\mbox.\egroup
  }{2013}]{DBLP:conf/nips/2013}
Burges, C. J.~C.; Bottou, L.; Ghahramani, Z.; and Weinberger, K.~Q., eds.
\newblock 2013.
\newblock {\em Advances in Neural Information Processing Systems 26: 27th
  Annual Conference on Neural Information Processing Systems 2013. Proceedings
  of a meeting held December 5-8, 2013, Lake Tahoe, Nevada, United States}.

\bibitem[\protect\citeauthoryear{Cai and Wang}{2018}]{KBGAN:2018}
Cai, L., and Wang, W.~Y.
\newblock 2018.
\newblock {KBGAN:} adversarial learning for knowledge graph embeddings.
\newblock In {\em Proceedings of the 2018 Conference of the North American
  Chapter of the Association for Computational Linguistics: Human Language
  Technologies, {NAACL-HLT} 2018, New Orleans, Louisiana, USA, June 1-6, 2018,
  Volume 1 (Long Papers)},  1470--1480.

\bibitem[\protect\citeauthoryear{{Cai}, {Zheng}, and
  {Chang}}{2018}]{tkde:survey}
{Cai}, H.; {Zheng}, V.~W.; and {Chang}, K.~C.
\newblock 2018.
\newblock A comprehensive survey of graph embedding: Problems, techniques, and
  applications.
\newblock {\em IEEE Transactions on Knowledge and Data Engineering}
  30(9):1616--1637.

\bibitem[\protect\citeauthoryear{Ebisu and
  Ichise}{2018}]{DBLP:conf/aaai/EbisuI18}
Ebisu, T., and Ichise, R.
\newblock 2018.
\newblock Toruse: Knowledge graph embedding on a lie group.
\newblock In McIlraith and Weinberger \shortcite{DBLP:conf/aaai/2018},
  1819--1826.

\bibitem[\protect\citeauthoryear{Fan \bgroup et al\mbox.\egroup
  }{2014}]{DBLP:conf/paclic/FanZCZ14}
Fan, M.; Zhou, Q.; Chang, E.; and Zheng, T.~F.
\newblock 2014.
\newblock Transition-based knowledge graph embedding with relational mapping
  properties.
\newblock In Aroonmanakun, W.; Boonkwan, P.; and Supnithi, T., eds., {\em
  Proceedings of the 28th Pacific Asia Conference on Language, Information and
  Computation, {PACLIC} 28, Cape Panwa Hotel, Phuket, Thailand, December 12-14,
  2014},  328--337.
\newblock The {PACLIC} 28 Organizing Committee and {PACLIC} Steering Committee
  / {ACL} / Department of Linguistics, Faculty of Arts, Chulalongkorn
  University.

\bibitem[\protect\citeauthoryear{Glorot \bgroup et al\mbox.\egroup
  }{2013}]{DBLP:journals/corr/abs-1301-3485}
Glorot, X.; Bordes, A.; Weston, J.; and Bengio, Y.
\newblock 2013.
\newblock A semantic matching energy function for learning with
  multi-relational data.
\newblock In Bengio, Y., and LeCun, Y., eds., {\em 1st International Conference
  on Learning Representations, {ICLR} 2013, Scottsdale, Arizona, USA, May 2-4,
  2013, Workshop Track Proceedings}.

\bibitem[\protect\citeauthoryear{Guo \bgroup et al\mbox.\egroup
  }{2017}]{DBLP:journals/tkde/GuoWWWG17}
Guo, S.; Wang, Q.; Wang, B.; Wang, L.; and Guo, L.
\newblock 2017.
\newblock {SSE:} semantically smooth embedding for knowledge graphs.
\newblock {\em {IEEE} Trans. Knowl. Data Eng.} 29(4):884--897.

\bibitem[\protect\citeauthoryear{He \bgroup et al\mbox.\egroup
  }{2015}]{DBLP:conf/cikm/HeLJ015}
He, S.; Liu, K.; Ji, G.; and Zhao, J.
\newblock 2015.
\newblock Learning to represent knowledge graphs with gaussian embedding.
\newblock In Bailey, J.; Moffat, A.; Aggarwal, C.~C.; de~Rijke, M.; Kumar, R.;
  Murdock, V.; Sellis, T.~K.; and Yu, J.~X., eds., {\em Proceedings of the 24th
  {ACM} International Conference on Information and Knowledge Management,
  {CIKM} 2015, Melbourne, VIC, Australia, October 19 - 23, 2015},  623--632.
\newblock {ACM}.

\bibitem[\protect\citeauthoryear{Jenatton \bgroup et al\mbox.\egroup
  }{2012}]{DBLP:conf/nips/JenattonRBO12}
Jenatton, R.; Roux, N.~L.; Bordes, A.; and Obozinski, G.
\newblock 2012.
\newblock A latent factor model for highly multi-relational data.
\newblock In Bartlett, P.~L.; Pereira, F. C.~N.; Burges, C. J.~C.; Bottou, L.;
  and Weinberger, K.~Q., eds., {\em Advances in Neural Information Processing
  Systems 25: 26th Annual Conference on Neural Information Processing Systems
  2012. Proceedings of a meeting held December 3-6, 2012, Lake Tahoe, Nevada,
  United States.},  3176--3184.

\bibitem[\protect\citeauthoryear{Ji \bgroup et al\mbox.\egroup
  }{2015}]{DBLP:conf/acl/JiHXL015}
Ji, G.; He, S.; Xu, L.; Liu, K.; and Zhao, J.
\newblock 2015.
\newblock Knowledge graph embedding via dynamic mapping matrix.
\newblock In {\em Proceedings of the 53rd Annual Meeting of the Association for
  Computational Linguistics and the 7th International Joint Conference on
  Natural Language Processing of the Asian Federation of Natural Language
  Processing, {ACL} 2015, July 26-31, 2015, Beijing, China, Volume 1: Long
  Papers},  687--696.
\newblock The Association for Computer Linguistics.

\bibitem[\protect\citeauthoryear{Ji \bgroup et al\mbox.\egroup
  }{2016}]{DBLP:conf/aaai/JiLH016}
Ji, G.; Liu, K.; He, S.; and Zhao, J.
\newblock 2016.
\newblock Knowledge graph completion with adaptive sparse transfer matrix.
\newblock In Schuurmans, D., and Wellman, M.~P., eds., {\em Proceedings of the
  Thirtieth {AAAI} Conference on Artificial Intelligence, February 12-17, 2016,
  Phoenix, Arizona, {USA.}},  985--991.
\newblock {AAAI} Press.

\bibitem[\protect\citeauthoryear{Lin \bgroup et al\mbox.\egroup
  }{2015a}]{PTransE}
Lin, Y.; Liu, Z.; Luan, H.; Sun, M.; Rao, S.; and Liu, S.
\newblock 2015a.
\newblock Modeling relation paths for representation learning of knowledge
  bases.
\newblock In M{\`{a}}rquez, L.; Callison{-}Burch, C.; Su, J.; Pighin, D.; and
  Marton, Y., eds., {\em Proceedings of the 2015 Conference on Empirical
  Methods in Natural Language Processing, {EMNLP} 2015, Lisbon, Portugal,
  September 17-21, 2015},  705--714.
\newblock The Association for Computational Linguistics.

\bibitem[\protect\citeauthoryear{Lin \bgroup et al\mbox.\egroup
  }{2015b}]{DBLP:conf/aaai/LinLSLZ15}
Lin, Y.; Liu, Z.; Sun, M.; Liu, Y.; and Zhu, X.
\newblock 2015b.
\newblock Learning entity and relation embeddings for knowledge graph
  completion.
\newblock In Bonet, B., and Koenig, S., eds., {\em Proceedings of the
  Twenty-Ninth {AAAI} Conference on Artificial Intelligence, January 25-30,
  2015, Austin, Texas, {USA.}},  2181--2187.
\newblock {AAAI} Press.

\bibitem[\protect\citeauthoryear{McIlraith and
  Weinberger}{2018}]{DBLP:conf/aaai/2018}
McIlraith, S.~A., and Weinberger, K.~Q., eds.
\newblock 2018.
\newblock {\em Proceedings of the Thirty-Second {AAAI} Conference on Artificial
  Intelligence, (AAAI-18), the 30th innovative Applications of Artificial
  Intelligence (IAAI-18), and the 8th {AAAI} Symposium on Educational Advances
  in Artificial Intelligence (EAAI-18), New Orleans, Louisiana, USA, February
  2-7, 2018}. {AAAI} Press.

\bibitem[\protect\citeauthoryear{Nguyen \bgroup et al\mbox.\egroup
  }{2018}]{DBLP:conf/naacl/NguyenNNP18}
Nguyen, D.~Q.; Nguyen, T.~D.; Nguyen, D.~Q.; and Phung, D.~Q.
\newblock 2018.
\newblock A novel embedding model for knowledge base completion based on
  convolutional neural network.
\newblock In Walker, M.~A.; Ji, H.; and Stent, A., eds., {\em Proceedings of
  the 2018 Conference of the North American Chapter of the Association for
  Computational Linguistics: Human Language Technologies, NAACL-HLT, New
  Orleans, Louisiana, USA, June 1-6, 2018, Volume 2 (Short Papers)},  327--333.
\newblock Association for Computational Linguistics.

\bibitem[\protect\citeauthoryear{Nickel, Tresp, and
  Kriegel}{2011}]{DBLP:conf/icml/NickelTK11}
Nickel, M.; Tresp, V.; and Kriegel, H.
\newblock 2011.
\newblock A three-way model for collective learning on multi-relational data.
\newblock In Getoor, L., and Scheffer, T., eds., {\em Proceedings of the 28th
  International Conference on Machine Learning, {ICML} 2011, Bellevue,
  Washington, USA, June 28 - July 2, 2011},  809--816.
\newblock Omnipress.

\bibitem[\protect\citeauthoryear{Pennington, Socher, and
  Manning}{2014}]{pennington2014glove}
Pennington, J.; Socher, R.; and Manning, C.~D.
\newblock 2014.
\newblock Glove: Global vectors for word representation.
\newblock In {\em Empirical Methods in Natural Language Processing (EMNLP)},
  1532--1543.

\bibitem[\protect\citeauthoryear{Socher \bgroup et al\mbox.\egroup
  }{2013}]{DBLP:conf/nips/SocherCMN13}
Socher, R.; Chen, D.; Manning, C.~D.; and Ng, A.~Y.
\newblock 2013.
\newblock Reasoning with neural tensor networks for knowledge base completion.
\newblock In Burges et~al. \shortcite{DBLP:conf/nips/2013},  926--934.

\bibitem[\protect\citeauthoryear{Trouillon \bgroup et al\mbox.\egroup
  }{2016}]{DBLP:conf/icml/TrouillonWRGB16}
Trouillon, T.; Welbl, J.; Riedel, S.; Gaussier, {\'{E}}.; and Bouchard, G.
\newblock 2016.
\newblock Complex embeddings for simple link prediction.
\newblock In Balcan, M., and Weinberger, K.~Q., eds., {\em Proceedings of the
  33nd International Conference on Machine Learning, {ICML} 2016, New York
  City, NY, USA, June 19-24, 2016}, volume~48 of {\em {JMLR} Workshop and
  Conference Proceedings},  2071--2080.
\newblock JMLR.org.

\bibitem[\protect\citeauthoryear{Trouillon \bgroup et al\mbox.\egroup
  }{2017}]{Trouillon:2017:JMLR}
Trouillon, T.; Dance, C.~R.; Gaussier, E.; Welbl, J.; Riedel, S.; and Bouchard,
  G.
\newblock 2017.
\newblock Knowledge graph completion via complex tensor factorization.
\newblock {\em Journal of Machine Learning Research} 18(1):4735--4772.

\bibitem[\protect\citeauthoryear{Wang \bgroup et al\mbox.\egroup
  }{2014}]{DBLP:conf/aaai/WangZFC14}
Wang, Z.; Zhang, J.; Feng, J.; and Chen, Z.
\newblock 2014.
\newblock Knowledge graph embedding by translating on hyperplanes.
\newblock In Brodley, C.~E., and Stone, P., eds., {\em Proceedings of the
  Twenty-Eighth {AAAI} Conference on Artificial Intelligence, July 27 -31,
  2014, Qu{\'{e}}bec City, Qu{\'{e}}bec, Canada.},  1112--1119.
\newblock {AAAI} Press.

\bibitem[\protect\citeauthoryear{Wang, Li, and
  Pan}{2018}]{DBLP:conf/aaai/WangLP18}
Wang, P.; Li, S.; and Pan, R.
\newblock 2018.
\newblock Incorporating {GAN} for negative sampling in knowledge representation
  learning.
\newblock In McIlraith and Weinberger \shortcite{DBLP:conf/aaai/2018},
  2005--2012.

\bibitem[\protect\citeauthoryear{Wang, Wang, and
  Guo}{2015}]{DBLP:conf/ijcai/WangWG15}
Wang, Q.; Wang, B.; and Guo, L.
\newblock 2015.
\newblock Knowledge base completion using embeddings and rules.
\newblock In Yang, Q., and Wooldridge, M.~J., eds., {\em Proceedings of the
  Twenty-Fourth International Joint Conference on Artificial Intelligence,
  {IJCAI} 2015, Buenos Aires, Argentina, July 25-31, 2015},  1859--1866.
\newblock {AAAI} Press.

\bibitem[\protect\citeauthoryear{Xiao \bgroup et al\mbox.\egroup
  }{2015}]{DBLP:journals/corr/0005HHZ15}
Xiao, H.; Huang, M.; Hao, Y.; and Zhu, X.
\newblock 2015.
\newblock Transg : {A} generative mixture model for knowledge graph embedding.
\newblock {\em CoRR} abs/1509.05488.

\bibitem[\protect\citeauthoryear{Xiao, Huang, and
  Zhu}{2016}]{DBLP:conf/ijcai/0005HZ16}
Xiao, H.; Huang, M.; and Zhu, X.
\newblock 2016.
\newblock From one point to a manifold: Knowledge graph embedding for precise
  link prediction.
\newblock In Kambhampati, S., ed., {\em Proceedings of the Twenty-Fifth
  International Joint Conference on Artificial Intelligence, {IJCAI} 2016, New
  York, NY, USA, 9-15 July 2016},  1315--1321.
\newblock {IJCAI/AAAI} Press.

\bibitem[\protect\citeauthoryear{Yang \bgroup et al\mbox.\egroup
  }{2015}]{DBLP:journals/corr/YangYHGD14a}
Yang, B.; Yih, W.; He, X.; Gao, J.; and Deng, L.
\newblock 2015.
\newblock Embedding entities and relations for learning and inference in
  knowledge bases.
\newblock In Bengio, Y., and LeCun, Y., eds., {\em 3rd International Conference
  on Learning Representations, {ICLR} 2015, San Diego, CA, USA, May 7-9, 2015,
  Conference Track Proceedings}.

\end{thebibliography}
\end{document}